\documentclass[10pt,twocolumn,letterpaper]{article}

\usepackage{cvpr}              
\usepackage[accsupp]{axessibility} 
\usepackage{xcolor} 
\usepackage{multirow}

\definecolor{cvprblue}{rgb}{0.21,0.49,0.74}

\newcommand{\corrauthor}[1]{\thanks{Corresponding author \dag Project Lead \\
The work was done during an internship at ARC Lab.
#1}}
\newcommand*\samethanks[1][\value{footnote}]{\footnotemark[#1]}
\usepackage[pagebackref,breaklinks,colorlinks,allcolors=cvprblue]{hyperref}

\def\paperID{5987} 
\def\confName{CVPR}
\def\confYear{2025}

\title{Mono2Stereo: A Benchmark and Empirical Study for Stereo Conversion}

\author{Songsong Yu\textsuperscript{1,2}, Yuxin Chen\textsuperscript{2}\dag, Zhongang Qi\textsuperscript{2}\corrauthor{}, Zeke Xie\textsuperscript{3}, Yifan Wang\textsuperscript{1},\\ 
Lijun Wang\textsuperscript{1}\samethanks{}, 
 Ying Shan\textsuperscript{2}, Huchuan Lu\textsuperscript{1}\\
DLUT\textsuperscript{1}, ARC Lab, Tencent PCG\textsuperscript{2}, HKUST(GZ)\textsuperscript{3}\\
}

\begin{document}
\maketitle
\begin{abstract}
With the rapid proliferation of 3D devices and the shortage of 3D content, stereo conversion is attracting increasing attention. Recent works introduce pretrained Diffusion Models (DMs) into this task. However, due to the scarcity of large-scale training data and comprehensive benchmarks, the optimal methodologies for employing DMs in stereo conversion and the accurate evaluation of stereo effects remain largely unexplored. In this work, we introduce the Mono2Stereo dataset, providing high-quality training data and benchmark to support in-depth exploration of stereo conversion. With this dataset, we conduct an empirical study that yields two primary findings. 1) The differences between the left and right views are subtle, yet existing metrics consider overall pixels, failing to concentrate on regions critical to stereo effects. 2) Mainstream methods adopt either one-stage left-to-right generation or warp-and-inpaint pipeline, facing challenges of degraded stereo effect and image distortion respectively. Based on these findings, we introduce a new evaluation metric, Stereo Intersection-over-Union, which prioritizes disparity and achieves a high correlation with human judgments on stereo effect. Moreover, we propose a strong baseline model, harmonizing the stereo effect and image quality simultaneously, and notably surpassing current mainstream methods. Our code and data will be open-sourced to promote further research in stereo conversion. Our models are available at \underline{\textcolor{red}{mono2stereo-bench.github.io}}.

\end{abstract}  

\section{Introduction}
Stereo conversion aims to convert monocular images or videos into content with stereoscopic effects. With the rapid advancement of Augmented Reality (AR) and Virtual Reality (VR) technologies, the demand for 3D content has substantially increased. This surge has consequently stimulated significant research interest in the field of stereo conversion. However, traditional approaches typically require manual depth mapping or binocular equipment, which are both time-consuming and costly, making it challenging to meet the growing demand for stereo content production.




In recent year, some works \cite{wang2024stereodiffusion, zhao2024stereocrafter} leverage deep generative models, especially pretrained diffusion models~\cite{rombach2022high, saharia2022photorealistic, song2020denoising, ho2020denoising, he2022latent, ho2022video, podell2023sdxl, brooks2024video}, for stereo conversion due to their superior image generation capabilities.
These works can be categorized into two-stage and single-stage approaches, depending on whether disparity estimation is required. Two-stage approaches~\cite{wang2024stereodiffusion, zhao2024stereocrafter} initially perform geometric warping of the left-view image based on a disparity map, and employ an inpainting model to eliminate the occlusion introduced by the warping operation at the second stage. 
These methods heavily depend on accurate depth or disparity estimation, often leading to noticeable artifacts in the generated outputs. Single-stage methods~\cite{xie2016deep3d}, in contrast, perform the conversion without explicit disparity estimation, directly generating the right-view image from left-view input. Given the subtle differences between the left and right view images, the one-stage approaches are prone to degenerating into identity mapping, resulting in insufficient stereoscopic effects.

Despite some efforts have been made to explore the use of DMs for stereo conversion, two critical issues remain largely unexplored due to the absence of large-scale, publicly available training datasets and a standardized benchmark. First, what are the relative merits of one-stage versus two-stage methods, and are there more effective approaches to utilizing DMs for stereo conversion? 
Second, how to accurately assess the stereoscopic effect of the generated content in a more reliable manner?


This work introduces Mono2Stereo, a new dataset comprising over 2.4 million stereo pairs across diverse scenes, including both indoor and outdoor environments, animation and real-life content, and scenes with varying complexity. Leveraging this dataset, we conduct a thorough empirical study, investigating two key issues and presenting two insightful findings. 1) 
Both mainstream approaches have their respective strengths and weaknesses. The single-stage paradigm excels in generating high-fidelity images, but its stereoscopic effects are less satisfactory. In contrast, two-stage models achieve better stereoscopic effects due to explicit viewpoint cues, yet they face challenges with compromised image quality.
2) Conventional metrics such as RMSE and SSIM assess all pixel discrepancies or image statistics but overlook the disparity relationships in stereo conversion tasks, thus failing to reflect the quality of stereoscopic effects accurately. 

Motivated by two key findings, we investigate the optimal strategies for leveraging diffusion models in stereo conversion. This exploration leads to the proposal of a strong dual-condition baseline model with Edge Consistency loss, which simultaneously achieves high-quality generation and superior stereo effects.
For dedicated stereoscopic quality metric, we design Stereo Intersection-over-Union (SIoU) based on carefully analyzing inter-view disparities and factors that significantly influence stereoscopic perception. Extensive human subjective evaluation shows that SIoU 
achieves high correlation with human judgments on stereo effects, complementing existing image quality metrics for a thorough stereo conversion evaluation.

In summary, our approach includes three main contributions:
\begin{itemize}

\item We construct Mono2Stereo, a large-scale benchmark designed for high-quality stereo conversion. This benchmark encompasses three key dimensions to facilitate a comprehensive evaluation of such methods.


\item We introduce Stereo Intersection-over-Union (SIoU), a novel and pioneering evaluation metric designed to assess the prominence of stereoscopic effects in stereo pairs. This metric effectively complements existing evaluation metrics for a thorough assessment.


\item Through extensive experiments, we establish a strong baseline model for stereo conversion. Benefiting from dual conditioning and Edge Consistency loss, our model achieves both compelling image quality and convincing stereo effects.



\end{itemize}

\section{Related Work}
\paragraph{2D to 3D Image Conversion.}
In the early stages of 2D to 3D image conversion, the relative depth of the target scene was designed manually, followed by horizontal shifting of the images. This process was highly labor-intensive. Eventually, fully automated methods for image conversion are developed, which can be broadly categorized based on whether they estimate disparity. The first category~\cite{zhang20113d, konrad2013learning, konrad20122d, wang2024stereodiffusion} involves a two-stage process: in the first stage, a depth map is estimated, which is then converted into disparity space to horizontally shift the image. The second stage renders the shifted image. These methods often require highly accurate depth estimation, leading many approaches to focus on improving this step~\cite{yang2024depth, ranftl2020towards, yu2024dme, ke2024repurposing}. However, they tend to produce poor-quality results due to cumulative errors from depth estimation and the need to adjust parameters such as baseline and focal length according to the scene range. The second category~\cite{xie2016deep3d} employs end-to-end methods that directly generate an alternate viewpoint image from the input image. Although these methods are simpler to operate, they face challenges due to the lack of sufficient stereo image datasets and degradation issues, where the model's output tends to resemble the input image more than the desired viewpoint. 
\vspace{-0.5cm}
\paragraph{Novel View Synthesis.}
Novel view synthesis (NVS) emerges as a popular area of research, with prominent examples including NeRF~\cite{mildenhall2021nerf, pumarola2021d, barron2022mip, xu2022sinnerf, jain2021putting, deng2023nerdi} and Gaussian Splatting~\cite{kerbl20233d}. This field categorizes into single-object NVS~\cite{chan2022efficient, gu2021stylenerf, chan2021pi, nguyen2019hologan, niemeyer2021giraffe, gao2022get3d} and multi-object NVS~\cite{xu2022sinnerf, skorokhodov20233d, sargent2023vq3d}. Single-object methods~\cite{melas2023realfusion, poole2022dreamfusion, chan2023generative, wang2023score, deng2023nerdi, liu2024one} often struggle with poor generalization and diversity, prompting ongoing studies~\cite{poole2022dreamfusion, lin2023magic3d, lorraine2023att3d, wang2024prolificdreamer, chen2023fantasia3d} aimed at improving the Score Distillation Sampling (SDS) algorithm. In contrast, multi-object scenarios present greater complexity due to intricate geometric relationships. For instance, Zero1to3~\cite{liu2023zero} utilizes camera extrinsic conditions to guide the generation of novel view images, while ZeroNVS~\cite{sargent2024zeronvs} employs six degrees of freedom in extrinsic modeling to explore NVS in complex scenes based on the SDS framework. The task of converting 2D images to 3D represents a specific subset of novel view synthesis within intricate scenes, where the camera extrinsic remains fixed to translation. Despite the seemingly minor changes in viewpoint, stereo conversion demands much higher precision in image alignment and consistency. 

\begin{table}[]
\centering
\setlength{\tabcolsep}{0.25em}
\fontsize{9pt}{10pt}\selectfont
\caption{Statistic of the different split of Mono2Stereo dataset across five categories. }   
\begin{tabular}{*{7}{c}}
  \toprule
   & {Indoor} & {Outdoor} & {Animation} & {Simple} & {Complex} \\
  \midrule
  Training set & 1,332K & 1,088K & 643K & 726K & 1,694K  \\
  Test set  &  500 & 500 & 500 & 500 & 500   \\
  \bottomrule
\end{tabular}
\label{tab:statistic}
\vspace{-0.45cm}
\end{table}

\vspace{-0.5cm}
\paragraph{Diffusion Models.}
Diffusion models serve as the foundation for image~\cite{rombach2022high, saharia2022photorealistic, song2020denoising, ho2020denoising} and video~\cite{he2022latent, ho2022video, podell2023sdxl, brooks2024video} generation tasks. These models, trained on web-scale data, possess strong geometric priors that attract attention for their application in discriminative tasks such as depth estimation and segmentation. Most relevant to our work is StereoDiffusion~\cite{wang2024stereodiffusion}, a training-free method that estimates image disparity and shifts the latent representation at specific denoising steps to generate corresponding stereo images. However, since this method does not utilize training data, it remains highly dependent on parameters and produce various artifacts in complex scenes. To address these issues, we collect over two million pairs of stereo images for training, leveraging a data-driven approach to achieve improved performance and reliability.

\section{Mono2Stereo Benchmark}
In this section, we introduce the Mono2Stereo dataset. \cref{med:dataset} elaborates on the dataset collection and partitioning. \cref{med:siou} presents our novel evaluation metric, SIoU.

\subsection{Dataset Construction}
\label{med:dataset}
\noindent \textbf{Data Collection.} High-quality, large-scale datasets are crucial for training robust stereo conversion models. However, there is a lack of publicly available datasets for this task currently. Therefore, we gather stereo content from the internet and conduct a thorough manual review to eliminate low-quality and potentially sensitive material. This process has resulted in the curation of 200 videos and films (over 20 million frames). 3D content are usually processed into various formats across different devices. To maximize information retention and allow for flexible conversions, we select a side-by-side 3D format, where each frame has corresponding left and right perspective images. To enhance dataset diversity and minimize inter-frame redundancy, we sample one frame among every eight frames. Finally, all frames were uniformly resized to a resolution of 960 $\times$ 540, yielding a final dataset size of 2.4 million pairs.
\noindent \textbf{Dataset Partitioning.} 
To thoroughly assess three critical aspects of stereo effects: disparity accuracy, scene complexity, and color consistency, we establish five distinct test categories.
As shown in \cref{tab:statistic}, we provide 500 non-overlapping test pairs for each category. Indoor and outdoor scenes are characterized by distinct disparity ranges. Simple and complex scene categories encompass a range of scene layouts and texture complexities.
Significant differences exist in the color distributions of animated content compared to natural images. By conducting evaluation within these categories, we can comprehensively assess the performance of stereo conversion models across different dimensions. Moreover, our benchmark can also be adapted to evaluate video stereo conversion. 



\subsection{Evaluation Methodology}
\label{med:siou}
\begin{figure}[t]
\centering
\includegraphics[width=0.98\linewidth]{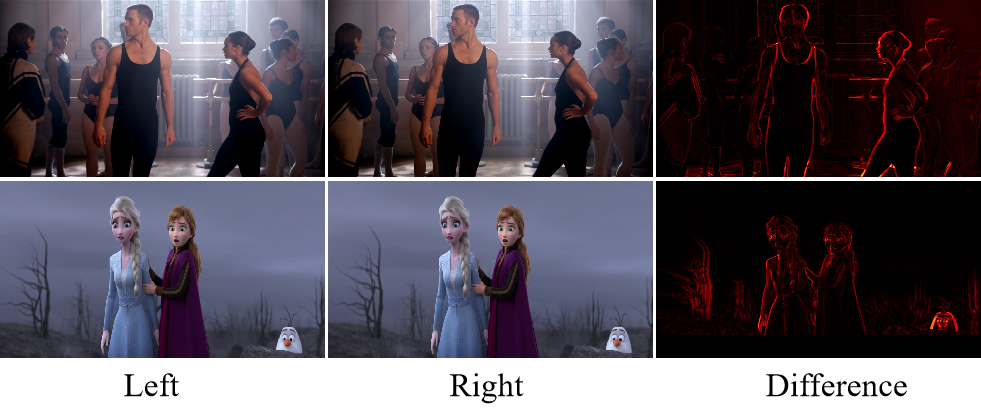}
\caption{Differences Visualization. Brighter (red) regions indicate larger differences between the left-view and right-view images, while darker regions represent smaller differences. Due to the small camera baseline, differences are primarily concentrated along object boundaries.}
\label{fig:diff}
\vspace{-0.3cm}
\end{figure}

Accurate evaluation metrics are cornerstones for the development of computer vision tasks. In the task of stereo conversion, traditional image generation metrics, 
\eg, RMSE, PSNR, and SSIM, are commonly adapted to evaluate the stereo conversion results. These metrics typically measure the global differences between generated results and ground truth to reflect the generation quality. However, as illustrated \cref{fig:diff}, the regions of difference between the left and right views, which are crucial for stereo effects, occupy only a small proportion of the entire image. Thus metrics that emphasize global differences may easily neglect the quality assessment of these crucial regions, making it challenging to accurately capture the stereoscopic quality of the generated results. We further conduct experimental validation on our dataset. As shown in \cref{tab:idm_compare}, we use RMSE, PSNR, SSIM as evaluation metrics and compare the performance of two models: a commercial stereo conversion model (OWL3D~\cite{owl3d}) and an identity mapping model whose output is also the input left-view image. It can be seen that although the identity mapping model generates results with no stereoscopic effect, it achieves high scores across all three metrics, further validating that these traditional metrics struggle to effectively evaluate stereoscopic quality.

\begin{table}[]
\centering
\setlength{\tabcolsep}{1.2em}
\fontsize{9pt}{10.2pt}\selectfont
\caption{Quantitative comparison of the OWL3D and identity mapping model on our Mono2Stereo dataset. \color{red}``IDM" \color{black}indicates the identity mapping model.}
\begin{tabular}{*{4}{c}}
  \toprule
  \textbf{Method} & RMSE$\downarrow$ & PSNR$\uparrow$ & SSIM$\uparrow$ \\
  \midrule
  \color{red}IDM & \color{red}5.07 & \color{red}34.76 & \color{red}0.788 \\
  OWL3D~\cite{owl3d} &  5.81 & 33.25 & 0.732 \\
  \bottomrule
\end{tabular}
\label{tab:idm_compare}
\vspace{-0.45cm}
\end{table}



In order to assess the stereoscopic quality of generated images accurately, we propose a novel evaluation metric, Stereo Intersection-over-Union (SIoU). Different from traditional metrics, we prioritize disparity consistency in our metric by ensuring that the difference between the generated image and the left-view image aligns with the difference between the ground truth right-view image and the left-view image. Furthermore, extensive observations reveal that distortions in edge generation during disparity shifts lead to an unconvincing stereo experience and degraded visual quality. Therefore, we aim to ensure that the edge structure of the generated image closely aligns with the ground truth. Based on these observations and considerations, we propose SIoU as a new metric that incorporates both disparity consistency and edge structure consistency. 
Denoting the left-view, right-view, and generated images as $\boldsymbol{X}_l$, $\boldsymbol{X}_r$ and $\boldsymbol{X}_r^0$ respectively, we first convert all images to their corresponding grayscale representations, $\boldsymbol{L}, \boldsymbol{R}, \boldsymbol{G}$, and formulate the SIoU as
\begin{equation}
\footnotesize
\begin{aligned}
    \text{SIoU}(\boldsymbol{L}, \boldsymbol{R}, \boldsymbol{G}) = \alpha \times \frac{\boldsymbol{E}_g \cap \boldsymbol{E}_r}{\boldsymbol{E}_g \cup \boldsymbol{E}_r} + (1-\alpha) \times \frac{|\boldsymbol{G}-\boldsymbol{L}| \cap |\boldsymbol{R}-\boldsymbol{L}|}{|\boldsymbol{G}-\boldsymbol{L}| \cup |\boldsymbol{R}-\boldsymbol{L}|},
\end{aligned}
\label{eq:siou}
\end{equation}
where $\boldsymbol{E}_r$ and $\boldsymbol{E}_g$ represent the single-channel binary edge maps extracted from $\boldsymbol{R}$ and $\boldsymbol{G}$ with the Canny~\cite{canny1986computational} edge detector. Specifically, the first item calculates the IoU between $\boldsymbol{E}_r$ and $\boldsymbol{E}_g$, quantifying the edge structure similarity between the generated image and the right view image. The second term leverages the differences between the left and right views as a reference and computes the IoU of the absolute difference maps, which reflects the disparity consistency of the generated image. It is important to note that, in the second term, the difference map between the two images is continuous. We binarize these maps by applying a threshold. In this work, we set the balancing parameter $\boldsymbol{\alpha}$ to 0.75. 
A comprehensive description of the experimental settings is available in the supplementary material.
 


\section{Diffusion Model for Stereo Conversion}
In this section, we conduct an empirical study on the training diffusion model of stereo conversion tasks. \cref{defination} introduces the problem definition of stereo conversion and the modeling approach using a diffusion model. In \cref{med:pipeline}, we investigate the impact of different conditions on the diffusion model. Finally, \cref{LEC loss} presents our proposed Edge Consistency loss to address the degradation issue.

\subsection{Problem Definition and Formulation}
\label{defination}
The stereo conversion task aims to generate a right-view image from the left-view image.
The one-stage paradigm takes the left-view image as input and estimates the distribution of right-view image directly, which is formulated as $D(\boldsymbol{X}_r | \boldsymbol{X}_l)$.
In a two-stage paradigm, depth or disparity is first estimated based on $\boldsymbol{X}_l$. Subsequently, the left-view image is warped based on the disparity map, resulting in $\boldsymbol{X}_w$, which serves as the input to the inpainting model. The two-stage paradigm can be formulated as $D(\boldsymbol{X}_r | \boldsymbol{X}_w)$. Both paradigms to conditional denoising problem share the same underlying model, differing only in their conditions.


To fully leverage the knowledge obtained from the pre-training of the diffusion model, we follow a successful practice, Marigold~\cite{ke2024repurposing}. By encoding RGB images into a low-dimensional latent space using a variational autoencoder $\boldsymbol{\varepsilon}: \boldsymbol{Z} = \boldsymbol{\varepsilon}(\boldsymbol{X})$, all denoising processes are conducted in this space, significantly reducing computational costs. Specifically, during the training phase, Gaussian noise ($\boldsymbol{\epsilon} \sim \mathcal{N}(\boldsymbol{0}, \boldsymbol{I})$) is gradually added to the latent presentation $\boldsymbol{Z}_r$ of $\boldsymbol{X}_r$ to obtain noisy samples $\boldsymbol{Z}_r^t$ as 
\begin{equation}
\begin{aligned}
    \boldsymbol{Z}^t_r = \sqrt{\bar{\boldsymbol{\alpha}}_t}\boldsymbol{Z}_r +  \sqrt{1- \bar{\boldsymbol{\alpha}}_t}\boldsymbol{\epsilon},
\end{aligned}
\label{noise eq}
\end{equation}
$\bar{\boldsymbol{\alpha}_t} := \prod^t_{j=1}(1-\boldsymbol{\beta}_j)$, where $t$ is the timestep and \{$\boldsymbol{\beta}_1, \dots, \boldsymbol{\beta}_j$\} is the variance schedule of a \textit{T} steps process. The conditional denoising model $\boldsymbol{\epsilon}_{\theta}$ progressively denoise $\boldsymbol{Z}_r^t$ to obtain $\boldsymbol{Z}_r^{t-1}$, where $\theta$ represents the learnable parameters. During the inference phase, a progressively cleaner estimate $\boldsymbol{Z}_r^0 = \boldsymbol{\epsilon}_\theta(\boldsymbol{Z}^t, \boldsymbol{Z}_l)$ is obtained by iteratively denoising from an initial normally-distributed variable $\boldsymbol{Z}^t$ through $t$ iterations of the denoiser $\boldsymbol{\epsilon}_\theta$. The $\boldsymbol{Z}_r^0$ is then decoded by the decoder to produce the prediction of the right-view image $\boldsymbol{X}_r^0$. It is worth noting that the decoding step is not required during the training process, thereby saving a considerable amount of time.

\subsection{Optimal Condition Exploration}
\label{med:pipeline}

\begin{figure}[t]
\centering
\includegraphics[width=0.475\textwidth]{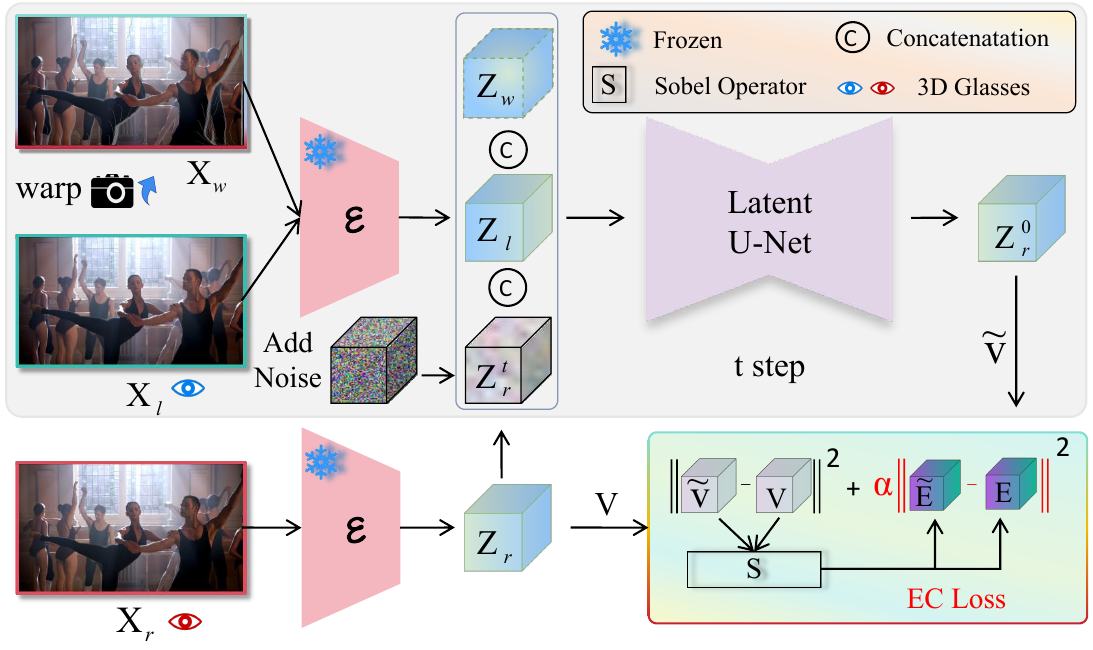}
\caption{Training pipeline of the dual-condition model. The images are fed into the VAE encoder to obtain the corresponding latent representations. The geometric and viewpoint conditions are concatenated as the input to the UNet, and only the Unet is optimized during training. To overcome degradation issues, additional constraints are applied to the edges of the velocity representations.}
\label{fig:training pipe}
\vspace{-0.5cm}
\end{figure}

For the architecture design, we first analyze the characteristics of the two paradigms. The input for the single-stage is the original left-view image $\boldsymbol{X}_l$, which provides complete structural and texture information. In contrast, two-stage paradigms utilize the disparity-warped image $\boldsymbol{X}_w$ as input, explicitly providing viewpoint information as guidance. For clarity, we define the latent representation of the left-view image as the geometric condition, denoted by  $\boldsymbol{Z}_l$. Meanwhile, the latent representation of the image resulting from the disparity transformation of the left-view is termed the viewpoint condition, $\boldsymbol{Z}_w$. 

To evaluate the performance of diffusion models under distinct paradigms, we assess each condition independently. Specifically, for the geometric condition, the left-view image is directly processed by a variational autoencoder (VAE). To derive the viewpoint condition, we first estimate the disparity map using a depth estimation model~\cite{yang2024depth}. The input image is then warped according to this disparity, and the resulting warped image $\boldsymbol{X}_w$ is input into the VAE to obtain $\boldsymbol{Z}_w$. Each condition is subsequently concatenated with the noisy sample $\boldsymbol{Z}_r^t$ of the right-view image and passed into the UNet for denoising.
Through experiments, we find that geometric condition yields superior image quality, while viewpoint conditioning produces more compelling stereo effects. Therefore, we explore a dual-condition approach, concatenating the geometric condition $\boldsymbol{Z}_l$, viewpoint condition $\boldsymbol{Z}_w$, and noisy right-view sample $\boldsymbol{Z}_r^t$ simultaneously as input, as demonstrated in \cref{fig:training pipe}. The objective is to equip the model with explicit viewpoint cues while simultaneously allowing it to reference complete geometric information, thereby facilitating the extraction of richer information from various input sources. To ensure compatibility with the original Latent Diffusion model~\cite{rombach2022high}, we duplicate the first convolutional layer of its UNet either once or twice, with the corresponding weights reduced by the appropriate factor. Details of the experiments are provided in \cref{exp:sota}. By employing dual-condition modeling, we demonstrate the feasibility of simultaneously achieving high image quality and compelling stereo effects.

\subsection{Analyzing and Mitigating Model Degradation}
\label{LEC loss}
In the task of stereo conversion, the overall difference between the left-view and right-view images is minimal, even in the latent space, with only a few elements exhibiting significant discrepancies. This presents a challenge for model optimization, as the input condition and the target to be estimated are highly similar, leading the model to take shortcuts. In other words, the images generated by the model tend to closely resemble the input images rather than the desired right-view images, as shown in row 2 of \cref{fig:vec}. At this point, the model almost degrades into an identity mapping, which is highly detrimental to the stereoscopic effects. It is worth noting that both paradigms exhibit varying degrees of degradation, with the issue being more pronounced for the single-stage geometric condition.

\begin{figure}[t]
\centering
\includegraphics[width=0.98\linewidth]{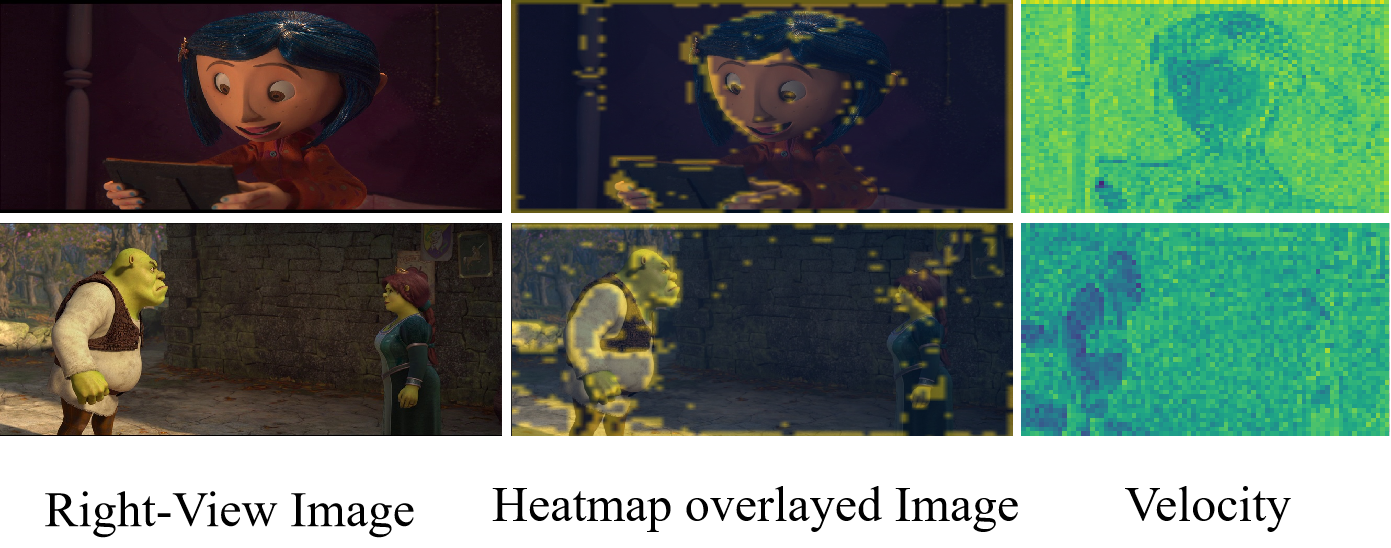}
\caption{Visualization of the optimization target. (Middle) Heatmap generated from ``Velocity'' using an edge detection operator, overlaid on the input image. (Right) The ``Velocity" exhibits a strong spatial correlation with the image content.}
\label{fig:velocity}
\vspace{-0.6cm}
\end{figure}

Our analysis of the stereo image pair, as depicted in \cref{fig:diff}, reveals that the main discrepancies between the left-view and the right-view images are centered around the high-frequency details at object edges. Therefore, a straightforward approach is to impose a consistency constraint that minimizes the distance between the edges of the estimated results $\boldsymbol{X}_r^0$ and the edges of the right-view image $\boldsymbol{X}_r$. However, directly optimizing the generated image incurs significant computational costs, and we aim to avoid forward propagation through the decoder during training. 
Several studies~\cite{wang2024stereodiffusion, ji2023ddp} highlight the spatial correlation between latent representations and images, an observation we also confirm in the context of the velocity prediction objective~\cite{salimans2022progressive}, as illustrated in \cref{fig:velocity}.
Leveraging this observation, we propose an Edge Consistency loss (EC loss). Specifically, we use Sobel edge extraction operators to enhance the edge information within the velocity representations $\boldsymbol{\epsilon}$ and $\boldsymbol{\hat{\epsilon}}$, and then impose constraints on the extracted edge information. We formulate Edge Consistency loss as follows, 
\begin{equation}
\label{eq:lec loss}
\begin{aligned}
\boldsymbol{\ell} =\mathbb{E}_{t,\boldsymbol{Z}_r,\boldsymbol{\epsilon}}\left \| \boldsymbol{\epsilon} -\hat{\boldsymbol{\epsilon}} \right \|^2 + \boldsymbol{\alpha} \mathbb{E} _{t,\boldsymbol{Z}_r,\boldsymbol{\epsilon}}\left \| S(\boldsymbol{\epsilon})- S(\hat{\boldsymbol{\epsilon}})\right \|^2,
\end{aligned}
\end{equation}
with $S$ representing the Sobel operator. To ensure that the values of the two loss terms are approximately the same, we set the balancing factor $\boldsymbol{\alpha}$ to 1. It is important to note that this constraint is applied universally, regardless of the specific paradigm.

\section{Experiments}

\begin{table*}[]
\centering
\setlength{\tabcolsep}{0.75em}
\fontsize{9pt}{10.2pt}\selectfont
\caption{Quantitative comparison of the proposed method across various conditions with SOTA methods. Mono2Stereo represents in-domain testing, while Inria 3DMovie represents out-of-domain testing. SIoU evaluates stereo quality, while RMSE and SSIM evaluate image quality. \color{red}``IDM'' \color{black} shows the comparison between the left and right views.}
\begin{tabular}{*{10}{c}}
  \toprule
  \multirow{2}{*}{\textbf{Method}} &\multicolumn{4}{c}{\textbf{Mono2Stereo}} &\multicolumn{4}{c}{\textbf{Inria 3DMovie}} &  \\
  \cmidrule(lr){2-5} \cmidrule(lr){6-9} 
   & SIoU$\uparrow$ & RMSE$\downarrow$ & PSNR$\uparrow$ & SSIM$\uparrow$ & SIoU$\uparrow$ & RMSE$\downarrow$ & PSNR$\uparrow$ & SSIM$\uparrow$ &\\
  \midrule
  \textcolor{red}{IDM} & \textcolor{red}{0.1644} & \textcolor{red}{5.07} & \textcolor{red}{34.76} & \textcolor{red}{0.788} & \textcolor{red}{0.1089} & \textcolor{red}{7.13} & \textcolor{red}{31.15} & \textcolor{red}{0.632} \\
  3D Photography~\cite{shih20203dphotography}  &  0.2103 & 8.89 & 29.30 & 0.285 & 0.2435 & 8.84 & 29.27 & 0.250 \\
  RePaint~\cite{lugmayr2022repaint}          & 0.1892 & 8.03 & 30.04 & 0.526 & 0.2391 & 8.85 & 29.20 & 0.335\\
  StereoDiffusion~\cite{wang2024stereodiffusion}  &  0.2375 & 7.42 & 30.96 & 0.621 & 0.2672 & 8.45 & 29.66 & 0.447\\
  OWL3D~\cite{owl3d} &  0.2781 & 5.81 & 33.25 & 0.732 & 0.2862 & 7.92 & 30.18 & 0.542 \\
  Geometric Condition & 0.2585 & \underline{5.40} & \underline{33.48} & \textbf{0.829} & 0.3135 & \underline{6.25} & \underline{32.21} & \textbf{0.767} \\
  Viewpoint Condition  & \underline{0.2809} & 6.12 & 32.53 & 0.762 & \underline{0.3381}& 6.66 & 31.66 & 0.760 \\
   Dual Condition  & \textbf{0.2836} & \textbf{5.34} & \textbf{34.09} & \underline{0.795} & \textbf{0.3398} & \textbf{6.11} & \textbf{32.55} & \underline{0.764} \\
  \bottomrule
\end{tabular}
\label{tab:sotas}
\vspace{-0.3cm}
\end{table*}







\vspace{-0.1cm}
In this section, we introduce our experimental setup in \cref{exp:setup}, detailing the implementation and the datasets. In \cref{exp:siou}, we present the user study of the proposed metric, SIoU, including comparisons with traditional metrics. In \cref{exp:sota}, we evaluate these metrics on several test datasets, showcasing the performance of various conditions and comparing them with other methods. Finally, we delve into a detailed discussion of the ablation experiments in \cref{exp:ablation}, including the impact of different conditions and the validation of the effectiveness of the Edge Consistency loss.

\subsection{Experimental Setup}
\label{exp:setup}
\paragraph{Implementation Details.}
Due to computational resource constraints, we opt to use an image generation model as our backbone, rather than a video generation model like our contemporaries in StereoCrafter~\cite{zhao2024stereocrafter}. During the training phase, we implement the DDPM noise scheduler~\cite{ho2020denoising} with 1000 diffusion steps, setting the initial learning rate to $1\times 10^{-5}$, and employing the warmup over the first 100 steps. Stable Diffusion V2~\cite{rombach2022high} serves as our initialization weights, and we employ Adam~\cite{kingma2014adam} as the optimizer. Our training setup comprises eight NVIDIA V100 GPUs, with a batch size of 48, over $150$ K iterations, taking approximately 7 days. In the testing phase, we follow the same approach as Marigold by using a DDIM schedule~\cite{song2020denoising} with 50 sampling steps. However, we do not need to perform inference repeatedly.

\begin{table}
\centering
\setlength{\tabcolsep}{7pt}
\fontsize{9pt}{10pt}\selectfont
\caption{Correlation with human judgements of stereo quality. SIoU outperforms other metrics in terms of both Spearman and Kendall rank correlation coefficients.}   
\begin{tabular}{*{6}{c}}
  \toprule
  \textbf{Metric}& {SIoU} & {RMSE} & {PSNR}& {SSIM} & \\
  \midrule
Spearman Rank& \textbf{0.84} &  0.26 & 0.26 & 0.21 &   \\
Kendall Rank & \textbf{0.73}  &  0.23 & 0.23 & 0.19  &  \\
  \bottomrule
\end{tabular} 
\label{tab:siou scores}
\vspace{-0.4cm}
\end{table}

\vspace{-0.54cm}
\paragraph{Datasets.}
We train our model using the Mono2Stereo dataset, which comprises $2.42$ million pairs of stereo images. Detailed information about the dataset is provided in \cref{med:dataset}. During training, the resolution is standardized to 640 pixels in width and 480 pixels in height. For testing, in addition to the five test subsets of our proposed dataset, we also collect stereo image pairs from the Inria 3DMovie dataset~\cite{seguin2014pose} to evaluate out-of-domain performance. The Inria 3DMovie dataset consists of 36 video clips, each containing between several dozen to over a hundred frames, totaling $2,727$ test pairs. In the testing phase, we compare several methods using our proposed SIoU metric as well as traditional metrics like RMSE, SSIM, and PSNR.  While RMSE, PSNR, and SSIM assess overall image quality, including aspects like brightness, contrast, and pixel-wise differences, SIoU specifically targets disparity perception and edge awareness to evaluate the stereoscopic quality of the generated images. To ensure fairness, we uniformly upsample the inference results to the original resolution of the images during testing.
\vspace{-0.1cm}
\subsection{Correlation with Human Judgements}
\label{exp:siou}
To evaluate the effectiveness of SIoU in assessing stereo quality, we conduct a human subjective evaluation using 1,100 side-by-side 3D image pairs generated by diffusion models under various conditions. These image pairs exhibit varying degrees of stereo fidelity, from a strong sense of immersion to a near-absence of stereo effects. Each test set comprises 5 to 10 stereo pairs of the same scene, generated by different models. Annotators are presented with these stereo pairs and the corresponding ground truth, and are asked to rate the perceived stereo quality on a scale of 1 to 10, with 1 representing the lowest and 10 the highest. To ensure objectivity, each test set is assessed by two annotators who are blinded to the source of the images. Annotators are instructed to first observe the entire set to ensure a consistent evaluation standard. For each sample, the scores from two annotators are averaged, which enables a correlation analysis with the relevant metrics.


\begin{figure*}[t]
\centering
\includegraphics[width=0.95\linewidth]{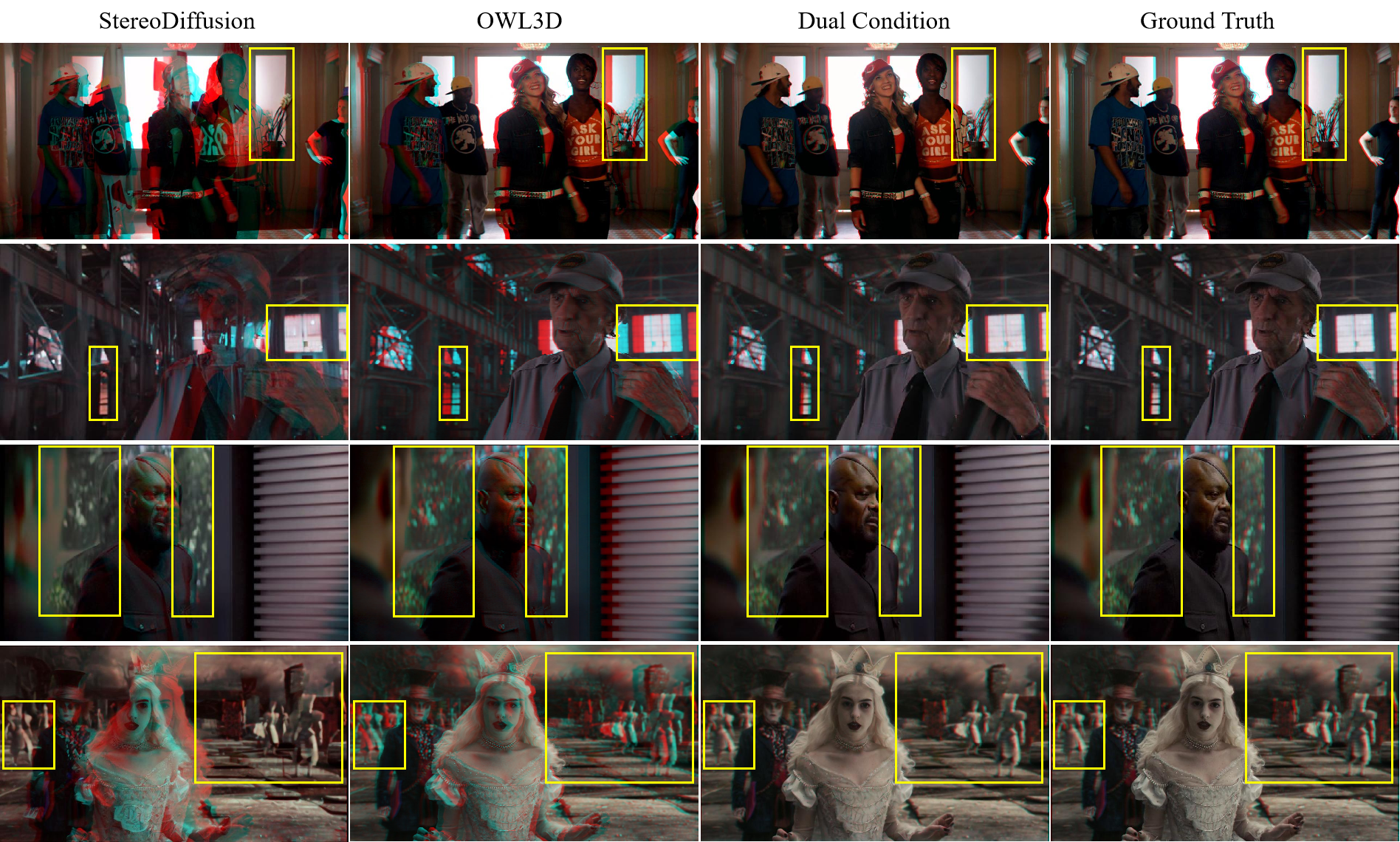}
\caption{Visual comparison of different methods using anaglyph (red-blue) stereo. StereoDiffusion and OWL3D exhibit artifacts such as unreasonable offsets,  while Dual Condition remains more faithful to the ground truth. The yellow boxes highlight the main differences.}
\label{fig:visual}
\vspace{-0.5cm}
\end{figure*}

We employ the Spearman and Kendall coefficients to evaluate the correlation between the metrics and the ranking of human assessments. Our evaluation reveals that three classic metrics, as shown in \cref{tab:siou scores}, fail to align well with human judgements. SSIM, a global statistic metric, and RMSE, based on pixel-wise differences, are inadequate at capturing the disparity information and edge differences essential for stereo perception. In contrast, SIoU attains the highest Spearman score of 0.84. The efficacy of SIoU in characterizing stereo quality stems from its emphasis on differences between left and right images, particularly at object edges. These factors are crucial cues for stereo effects. Therefore, SIoU can serve as a valuable complement to existing metrics.




\begin{table}
\centering
\setlength{\tabcolsep}{0.85em}
\fontsize{9pt}{10.2pt}\selectfont
\caption{Evaluation of the dual-condition model across five scenes.}    
   \begin{tabular}{*{5}{c}}
  \toprule
  \textbf{Scene}& {SIoU} & {RMSE} & {PSNR} & {SSIM} \\
  \midrule

Indoor  & {0.2819} & 5.21  & 34.12 & 0.815\\
Outdoor & {0.2969}  & 5.65 & 33.51 & 0.765 \\
Simple  & \textbf{0.3095}  & \textbf{4.29}  & \textbf{36.08} & \textbf{0.850} \\
Complex & {0.2894}  & 5.78  & 33.59 & 0.781\\
Animation& {0.2999}  & 5.76 & 33.13 & 0.762 \\

\bottomrule  
\end{tabular}

\label{tab:5settings-2con}
\vspace{-0.45cm}
\end{table}

\subsection{Evaluation on Mono2Stereo Benchmark}
\label{exp:sota}
We compare our approach against several state-of-the-art stereo conversion methods~\cite{lugmayr2022repaint, shih20203dphotography, wang2024stereodiffusion, owl3d}.
StereoDiffusion~\cite{wang2024stereodiffusion} is a training-free method that utilizes the prior knowledge of pretrained diffusion models by applying disparity transformations on the latent space to generate stereo images. 3D Photography~\cite{shih20203dphotography} and Repaint~\cite{lugmayr2022repaint}, two powerful inpainting models, are used to complete the missing regions in the warped images.
As shown in \cref{tab:sotas}, our method, evaluated under three different conditions, consistently outperforms most previous approaches in terms of both traditional measures and the proposed SIoU for stereo quality assessment. This highlights the significance of training data. Notably, single-stage geometric-conditioned diffusion models demonstrate superior image quality. However, their SIoU scores are limited by the absence of explicit disparity guidance. Conversely, two-stage viewpoint-conditioned diffusion models exhibit enhanced stereo consistency due to explicit viewpoint cues, but at the expense of image quality. Our proposed dual-condition paradigm effectively addresses this trade-off, achieving both high image quality and strong stereo effect.
\cref{fig:visual} illustrates that the dual-condition model produces anaglyph images most faithful to the ground truth. In contrast, StereoDiffusion and OWL3D exhibit excessive disparity in their anaglyph outputs, likely due to inaccurate disparity transformation, leading to a suboptimal visual experience. We also evaluate the performance of the dual-condition model across various scenes. As shown in \cref{tab:5settings-2con}, there is room for improvement in the stereo quality of complex and indoor scenes, as well as in the image quality of animated scenes. These findings indicate three crucial areas for future research: enhanced disparity accuracy, better geometric perception capabilities, and more advanced color consistency. 


\begin{table}[]
\centering
\setlength{\tabcolsep}{0.55em}
\fontsize{9pt}{10.2pt}\selectfont
\caption{Model impact under various conditions.}   
   \begin{tabular}{*{6}{c}}
  \toprule
  \multirow{2}{*}{\textbf{Conditions}} &\multicolumn{4}{c}{\textbf{Mono2Stereo}} &  \\
  \cmidrule(lr){2-6} 
   & SIoU$\uparrow$ & RMSE$\downarrow$ & PSNR$\uparrow$ & SSIM$\uparrow$ &\\
  \midrule
  Geo. Only & 0.2533 & \underline{6.97} & \underline{31.26} & \textbf{0.727} & \\
  Geo. + Depth & {0.2551} & 7.10 & 31.10 & 0.681 & \\
  Geo. + Edge  & 0.2543 & 7.20 & 30.98 & 0.675 & \\
  View. Only & 0.2604 & 7.15 & 31.04 & 0.695 & \\
  View. + Depth & \underline{0.2605} & 7.31 & 30.93 & 0.687 & \\
   View. + Edge  & 0.2596 & 7.42 & 30.72 & 0.683 &  \\
   view.+ Mask & 0.2581 & 7.40 & 30.75 & 0.672 \\
   Geo. + View.  & \textbf{0.2619} & \textbf{6.82} & \textbf{31.45} & \underline{0.721} & \\
   







  \bottomrule
\end{tabular} 
\label{tab:ablation on various con.}
\vspace{-0.35cm}
\end{table}

\subsection{Ablation Study}
\label{exp:ablation}
For ablation, we randomly select a consistent set of $180,000$ image pairs for training. We use the Mono2Stereo test set for in-domain validation and 36 video clips from the Inria 3DMovie dataset for out-of-domain testing. 
\vspace{-0.525cm}
\paragraph{Various Conditions.}
Due to the high dependency of the stereo transformation task on disparity and geometric information, we extensively explore the impact of different modes of conditioning on this task.
In addition to the viewpoint and geometric conditions outlined in \cref{exp:sota}, we investigate the use of depth maps, edge maps extracted from the left-side image, and occlusion masks, similar to those employed in StereoCrafter, as supplementary conditions. Intuitively, depth maps and edge information provide more direct parallax and texture cues. Specifically, depth maps are extracted using DepthAnything~\cite{yang2024depth}, and edges are derived using the Sobel operator. To adapt these inputs for the VAE encoder, we simply replicate them into three channels.
To avoid excessive model complexity and to save on training and inference time, we limit the input to a maximum of two conditions at a time. The experimental results, as shown in \cref{tab:ablation on various con.}, reveal that incorporating additional conditions does not consistently yield improvements comparable to the dual-condition framework and even has adverse effects, particularly on image quality metrics like PSNR. We hypothesize that this discrepancy stems from the significant distribution shift between inputs like edge maps, depth maps, and masks, compared to natural images. Since both the encoder and Unet models are pretrained on natural images, this gap hinders their ability to effectively leverage these supplementary cues. Therefore, exploring alternative alignment methods may be crucial for harnessing the full potential of these additional conditions.

\begin{table}[]
\centering
\setlength{\tabcolsep}{0.25em}
\fontsize{9pt}{10.2pt}\selectfont
\caption{Impact of EC loss across three conditions. EC loss consistently improves performance, with notable gains in SIoU, the metric for perceived stereo quality.}   
\begin{tabular}{*{8}{c}}
  \toprule
  \multirow{2}{*}{\textbf{Geo.}} & \multirow{2}{*}{\textbf{View.}} & \multirow{2}{*}{\textbf{LEC Loss}} &\multicolumn{4}{c}{\textbf{Mono2Stereo}} &  \\
  \cmidrule(lr){4-7} 
  & & & SIoU$\uparrow$ & RMSE$\downarrow$ & PSNR$\uparrow$ & SSIM$\uparrow$ &\\
  \midrule
  
  $\color{red}\checkmark$ &  &  & 0.2418 & 7.26 & 30.92 & 0.715  \\
  $\color{red}\checkmark$ & & $\color{red}\checkmark$ &  0.2533 & 6.97 & 31.26 & \textbf{0.727} \\
   & $\color{blue}\checkmark$  &  & 0.2573 & 7.37 & 30.78 & 0.661 & \\
   & $\color{blue}\checkmark$  & $\color{blue}\checkmark$  & 0.2604 & 7.15 & 31.04 & 0.695 \\
   $\color{green}{\checkmark}$ & $\color{green}\checkmark$ &  & 0.2588 & 6.90 & 31.35 & 0.721 \\
   $\color{green}\checkmark$ & $\color{green}\checkmark$ & $\color{green}\checkmark$ & \textbf{0.2619} & \textbf{6.82} & \textbf{31.45} & 0.721 \\
  \bottomrule

\end{tabular}




\label{tab:ablation on vec loss}
\vspace{-0.25cm}
\end{table}

\begin{figure}[t]
\centering
\includegraphics[width=0.95\linewidth]{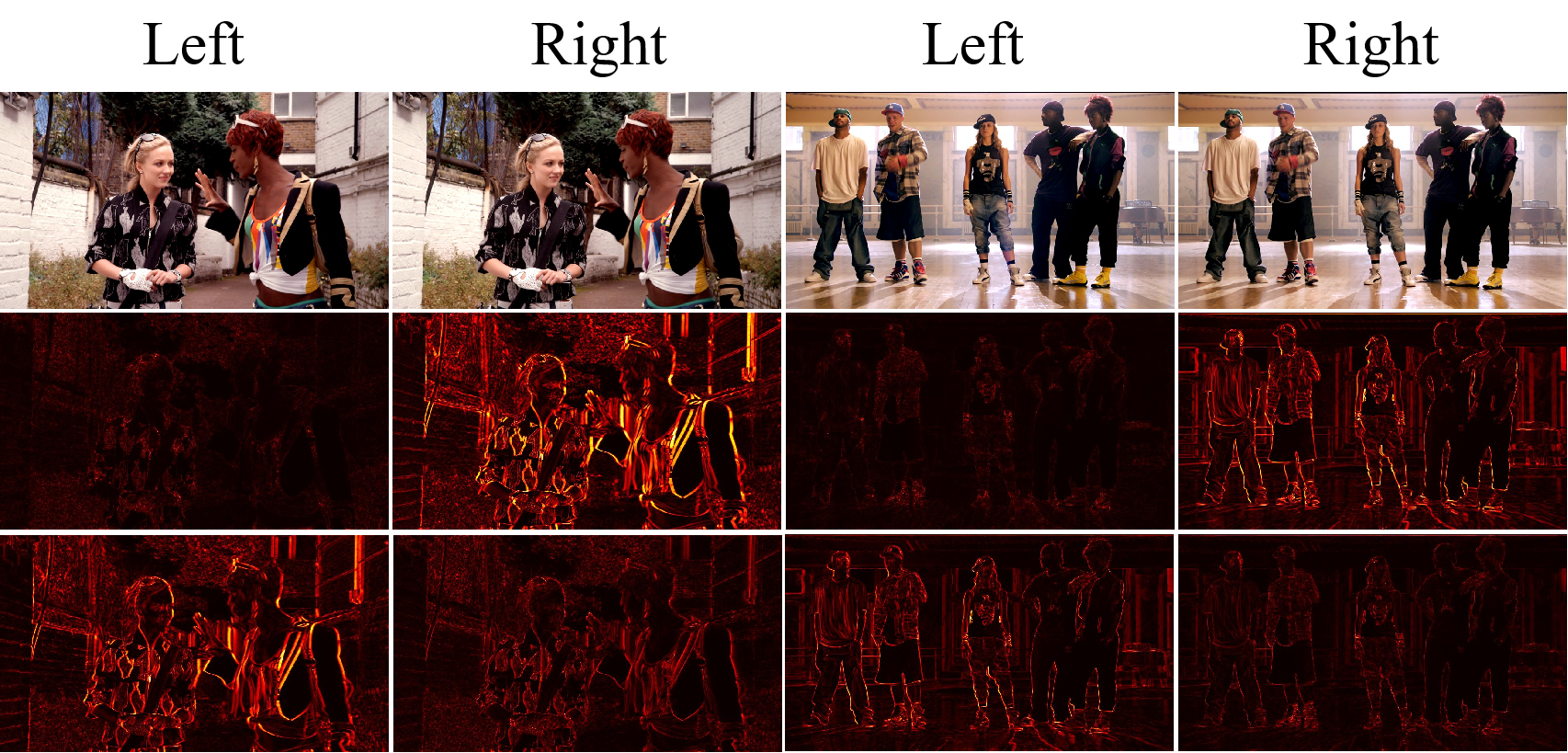}
\caption{Impact of EC loss on single-stage results under geometric constraints. (Row 2) Without EC loss, the estimation exhibits smaller differences with the left-view image but larger differences with the right-view image. (Row 3) EC loss significantly reduces the differences between the estimation and the right image.}
\label{fig:vec}
\vspace{-0.5cm}
\end{figure}

\vspace{-0.525cm}
\paragraph{Edge Consistency loss.}
The Edge Consistency loss encourages the model to prioritize consistency between the edges of the generated image and the ground truth throughout the optimization process. For single-stage models, this mitigation strategy helps prevent convergence to suboptimal solutions characterized by degradation. In two-stage models, it corrects for artifacts arising from inaccurate disparity estimation, such as blurred edges. Therefore, we evaluate the impact of applying the Edge Consistency loss to each condition described in \cref{med:pipeline}. The detailed results are shown in \cref{tab:ablation on vec loss}, indicating the greatest improvement under the geometric condition, with varying degrees of improvement observed for other conditions as well. This aligns with our initial observation that using only the left image as input leads to significant degradation in model performance. Additionally, in \cref{fig:vec}, we illustrate the impact of EC loss under the geometric condition, showing a marked reduction in the difference from the right image after introducing this loss term.

\section{Conclusion}
\vspace{-0.1cm}
This work introduces Mono2Stereo, a novel dataset designed to address the lack of publicly available data for stereo conversion tasks. Based on this dataset, we propose Stereo Intersection-over-Union (SIoU), a new evaluation metric for stereoscopic quality. Methodologically, we conduct empirical studies and introduce a robust, dual-conditional baseline framework that achieves high-quality generation with convincing stereo effects. Furthermore, we demonstrate that incorporating a Edge Consistency loss mitigates degradation and enhances stereo quality. We believe our benchmark will facilitate further advancements in stereoscopic conversion task.



\vspace{-0.5cm}
\paragraph{Acknowledgement.} This work is supported by the National Natural Science Foundation of China (62422610, U23A20386, 62276045, 62293542, 6244101029),  Liao Ning Province Science and Technology Plan (2023JH26, 10200016), Dalian City Science and Technology Innovation Fund (2023JJ11CG001),  the Ningbo Science and Technology Innovation Project (2024Z294), and the Fundamental Research Funds for the Central Universities (DUT22ZD210). 
{
    \small
    \bibliographystyle{ieeenat_fullname}
    \bibliography{main}
}


\clearpage  
\section{Appendix}
In this supplementary material, we provide detailed data processing methods and statistical details in Section~\ref{app:data}. Section~\ref{app:siou} elaborates on the parameter settings for SIoU. Additionally, Section~\ref{app:exp} presents supplementary experimental results.
\appendix
\section{Data Curation and Preprocessing.}
\label{app:data}

\begin{figure}[h]
\centering
\includegraphics[width=0.95\linewidth]{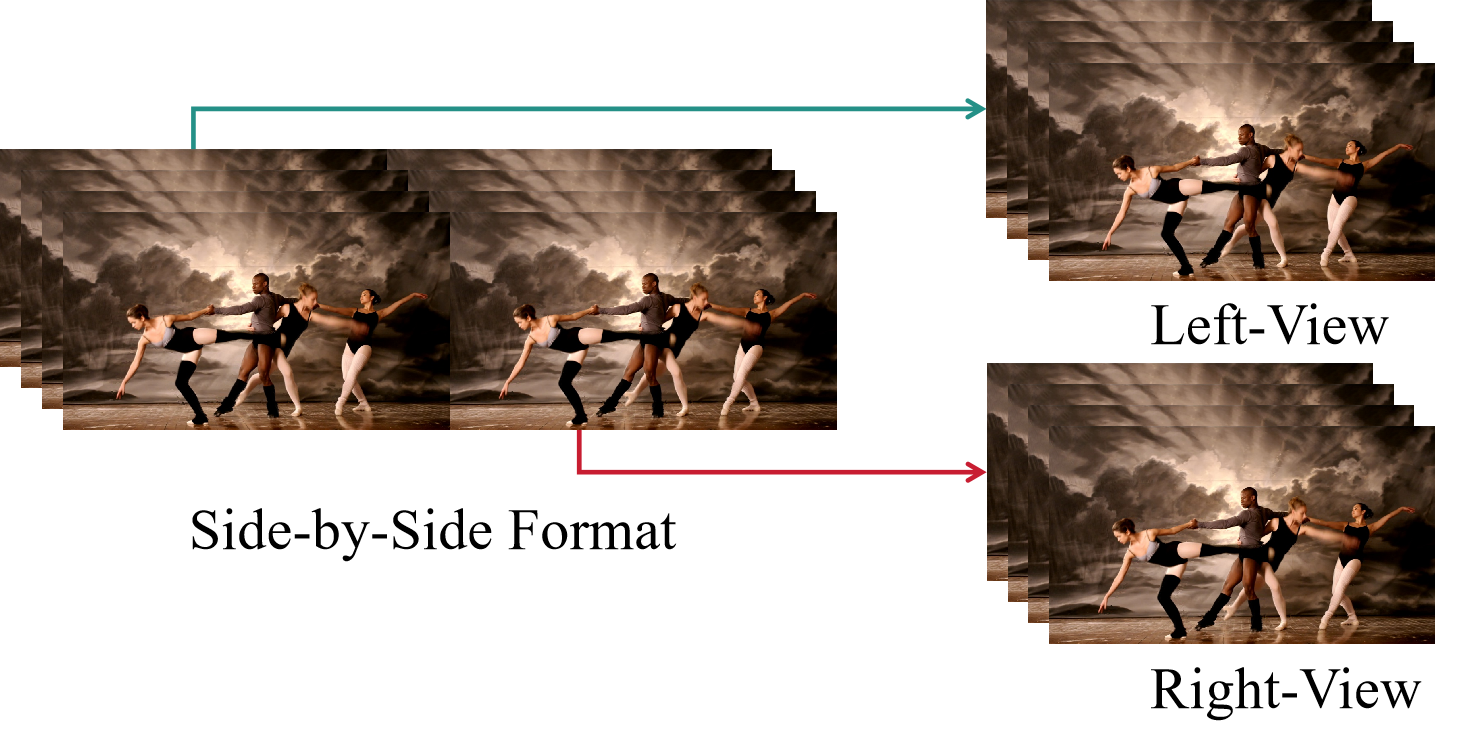}
\caption{Input (Left-View) and ground truth (Right-View) images produced by splitting a video frame.}
\label{fig:seg}
\end{figure}

We collect a substantial amount of 3D content in the left-right format from movies and videos, as illustrated in~\cref{fig:seg}. This format necessitates specific viewing equipment, such as 3D glasses, to ensure that the left and right eyes perceive the corresponding Left-View and Right-View images, respectively. By dividing these images from the middle, we create two distinct perspectives: the Leff-View and the Right-View images. Conversion to other stereoscopic formats can be achieved by applying appropriate processing techniques to this image pair.

\begin{figure}[h]
\centering
\includegraphics[width=\linewidth]{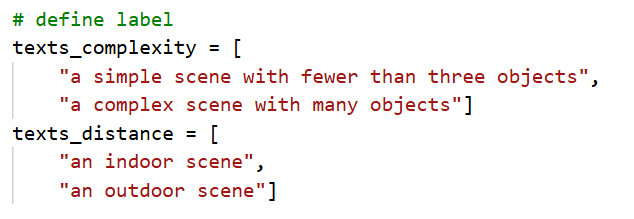}
\caption{CLIP scene categories.}
\label{fig:prompt}
\end{figure}

\begin{figure}[h]
\centering
\includegraphics[width=\linewidth]{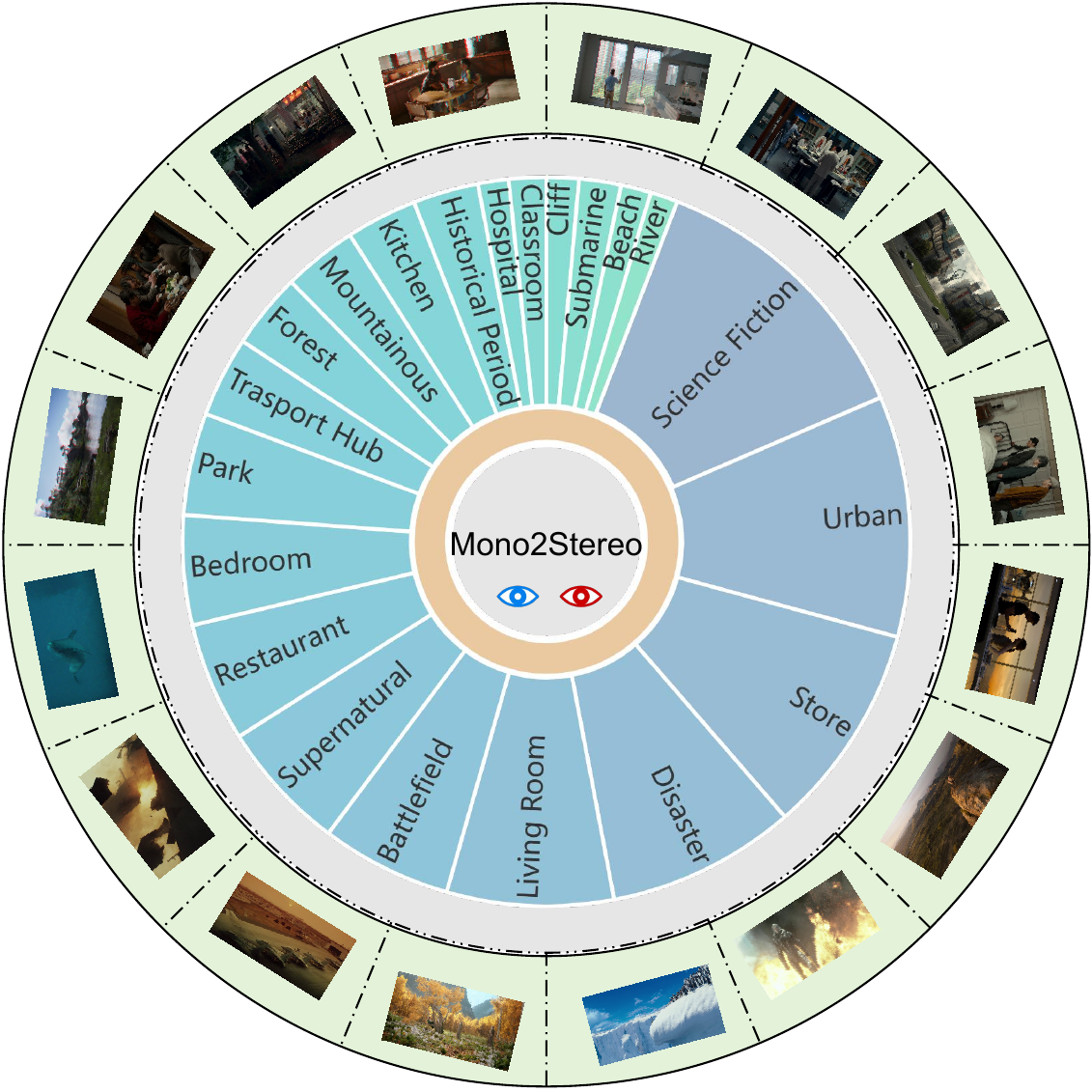}
\caption{Distribution Characteristics of the Mono2Stereo Dataset.}
\label{fig:distribution}
\end{figure}

For data statistics, we employ CLIP~\cite{clip} as a scene classifier. Specifically, we feed text prompts and images into text encoder and image encoder of CLIP, respectively. We then calculate the cosine similarity between the resulting embeddings, assigning the category with the highest similarity as the classification result. \cref{fig:prompt} showcases the specific text prompts used. We perform pairwise statistics for the four categories (indoor, outdoor, simple, and complex). Additionally, we analyze the scene distribution within the dataset. As illustrated in \cref{fig:distribution}, Mono2Stereo encompasses common indoor environments like living rooms and bedrooms, as well as more unique settings such as underwater scenes, cliffs, and rivers. For overall scene category statistics in ~\cref{fig:distribution}, we utilize prompts in the format of ``a/an [category] scene.''

\section{Parameter Settings for SIoU.}
\label{app:siou}
To illustrate the individual roles of the two terms within SIoU, we conduct separate human subjective evaluations, with the results presented in \cref{tab:siou terms}. As shown, both terms contribute to achieving good consistency, suggesting that each reflects stereo quality to a certain extent by primarily focusing on the true disparity regions between the Left-View and Right-View images. Regarding the balancing parameter $\boldsymbol{\alpha}$ in \cref{eq:siou}, we randomly divide 1100 image pairs into two sets: 500 pairs for optimal parameter and threshold searching, and 600 pairs for generalization validation. We experiment with various parameter settings for $\boldsymbol{\alpha}$, including 0.25, 0.5, 0.7, 0.75, and 0.8. Our findings indicate that these settings yield better consistency compared to a single item. Notably, $\boldsymbol{\alpha} = 0.75$ demonstrates the highest level of consistency. Therefore, we set $\boldsymbol{\alpha}$ to 0.75 for final SIoU. Validation on a set of 600 pairs, as illustrated in \cref{tab:siou terms}, shows no significant signs of overfitting.

\begin{table}[t]
\centering
\setlength{\tabcolsep}{0.95em}
\fontsize{9pt}{9pt}\selectfont
\caption{Correlation with human judgements of stereo quality. IoU1 and IoU2 are components of the proposed SIoU metric. Both demonstrate correlation with human perception. Combining these components into SIoU yields even higher correlation scores. The results are based on a validation set of 600 pairs.}    
   \begin{tabular}{*{4}{c}}
  \toprule
  \textbf{Metric}& \textbf{SIoU} & \textbf{IoU1} & \textbf{IoU2}   \\
  \midrule
Spearman Rank& \textbf{0.84} & 0.81 & 0.80  \\
Kendall Rank & \textbf{0.73}  & 0.70  & 0.68  \\
  \bottomrule
\end{tabular}

\label{tab:siou terms}
\end{table}

\begin{figure}[h]
\centering
\includegraphics[width=\linewidth]{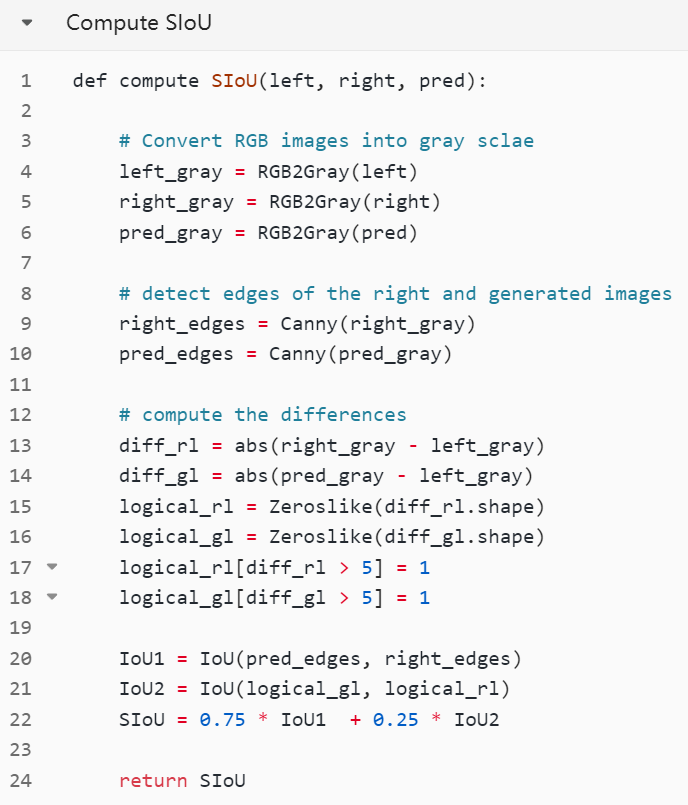}
\caption{Pseudocode for the SIoU calculation process.}
\label{fig:pseudocode for SIoU}
\end{figure}

For IoU2, employing a lower threshold ensures greater sensitivity to discrepancies, encompassing both disparity and pixel shifts. When validating across 500 sample pairs, we observe that a threshold of 5 yields the highest consistency. Attempting to decrease this threshold further actually reduces consistency. This occurs because a lower threshold ($ < 5$) incorporates more pixels into consideration, including those in areas that do not significantly impact the stereo effect, which is undesirable.


\section{Supplementary Experimental Results.}
\label{app:exp}
\subsection{Detailed Analysis of Two Conditions}
\begin{figure}[h]
\centering
\includegraphics[width=\linewidth]{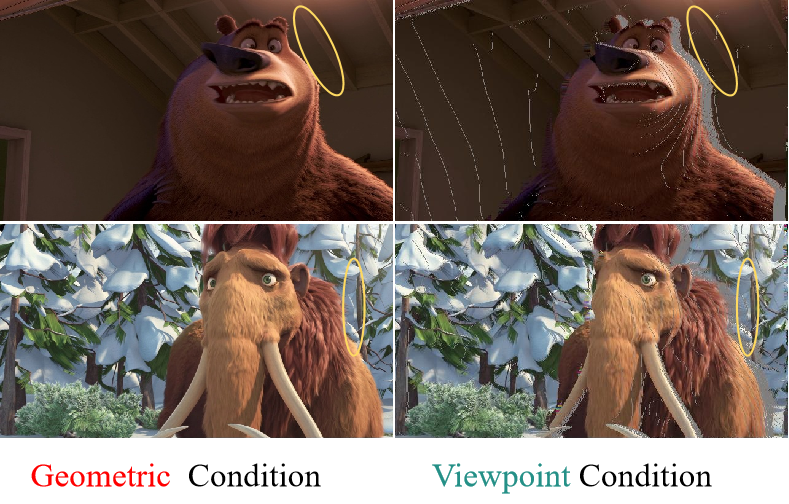}
\caption{Visualization of Dual-Condition. The yellow circles highlight the differences in key spatial relationships, while the gray areas represent geometric differences.}
\label{fig:dual-condition}
\end{figure}

In this paper, we define the complete Left-View image as the geometric condition, while the warped version of the Left-View image serves as the viewpoint condition. These conditions correspond to the inputs of single-stage and two-stage models, respectively. This section provides further clarification. As depicted in \cref{fig:dual-condition}, the Left-View is a complete natural image, offering comprehensive geometric structure and texture details. Conversely, the Warped image, derived from the Left-View image through disparity warping, exhibits a perspective closer to the Right-View image. Therefore, the Left-View image provides richer geometric information, while the Warped image explicitly offers an observational viewpoint, spatially aligning it closer to the target. This distinction forms the basis for our naming convention and motivates our design of the dual-condition model, leveraging the complementary strengths of both conditions.

\begin{figure}[h]
\centering
\includegraphics[width=\linewidth]{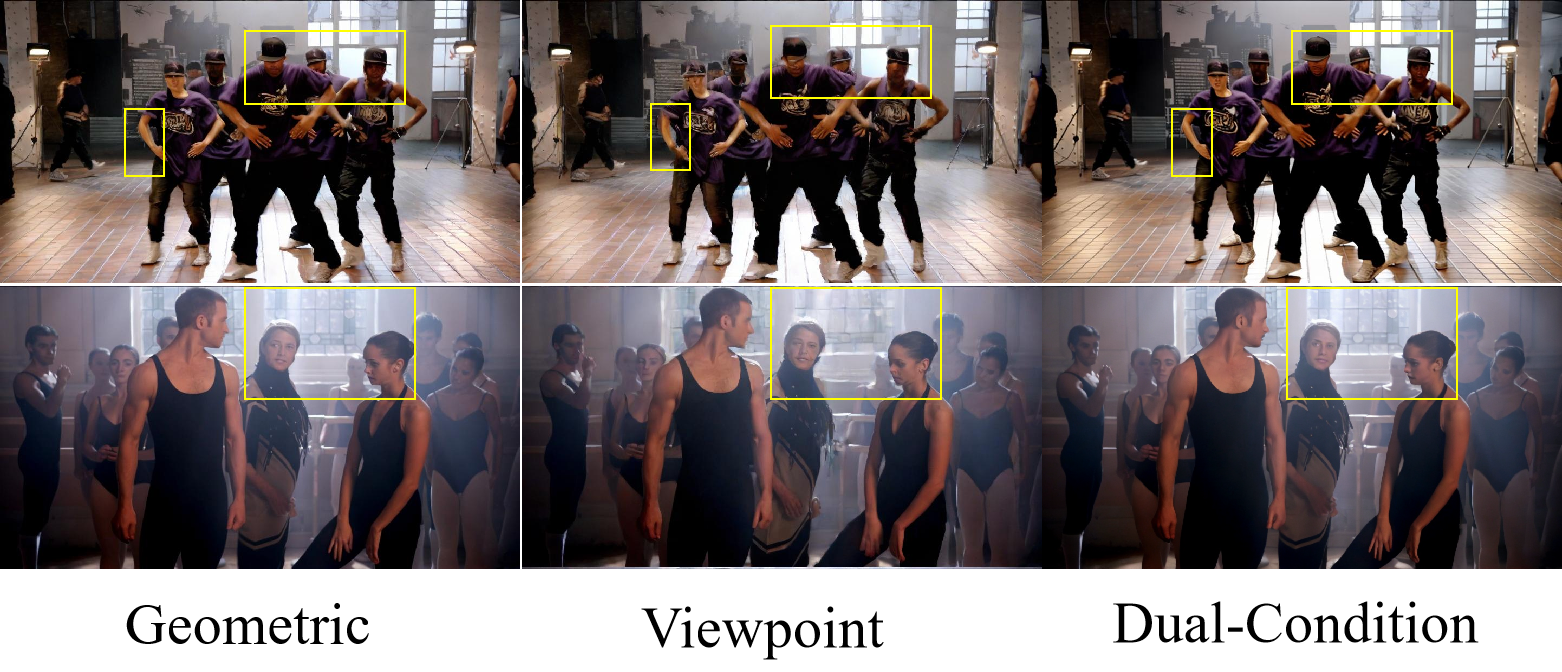}
\caption{The influence of identical conditions on the output results. Areas with significant differences are highlighted by yellow boxes.}
\label{fig:comparison with various con}
\end{figure}

Furthermore, we present results under three different conditions, as illustrated in \cref{fig:comparison with various con}. Both the ``Geometric'' and ``Viewpoint'' conditions exhibit artifacts to varying degrees, with the ``Viewpoint'' condition displaying more pronounced artifacts due to its partially occluded input. In contrast, the ``Dual-Condition'' yields superior image quality.

\subsection{Evaluating Performance in Various Scenes}
To gain a deeper understanding of the performance across different scenarios, we evaluate models separately on five distinct scenes from the Mono2Stereo test dataset. As shown in \cref{tab:5settings}, we observe that the model struggles in pairwise comparisons involving indoor, complex, and animation scenes. We hypothesize that this is due to limitations in three key areas where the model requires further improvement: disparity range estimation accuracy, geometric understanding, and color distribution handling. Consequently, we suggest that future research should focus on addressing these aspects. Finally, Mono2Stereo also provides 20 video clips for evaluating models. Despite our method being single-frame based, it still achieves promising results.


\subsection{Why Velocity Edges?}
\begin{figure}[h]
\centering
\includegraphics[width=\linewidth]{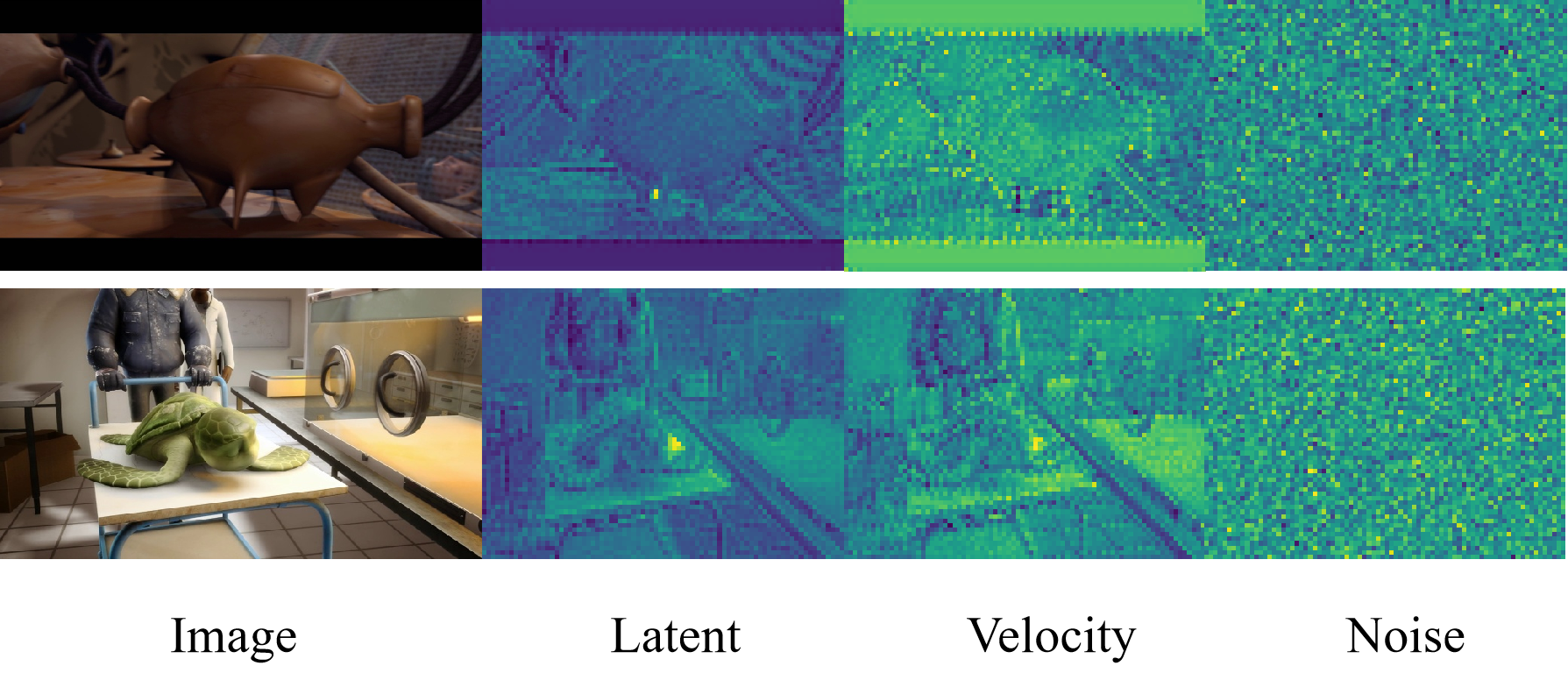}
\caption{Feature maps of latent and velocity.}
\label{fig:feat}
\end{figure}

Regarding the edge consistency constraint, the most intuitive approach appears to be constraining the edges within the latent space. Visualization of the feature maps, as illustrated in \cref{fig:feat}, confirms that both the latent and velocity exhibit positional correlation with the image. However, during training, we observe that predicting the latent or noise results in significantly slower convergence and even optimization failure, while velocity prediction does not suffer from these issues. Consequently, we opt to constrain the edges of the velocity field.

\subsection{Ablation Study on Inria 3DMovie.}
To further validate the effectiveness of the Edge Consistency loss, we conduct out-of-domain performance evaluations using the Inria 3DMovie dataset, which comprises $2,727$ stereoscopic image pairs. As shown in \cref{tab:ablation of lec loss on inria}, incorporating the Edge Consistency constraint consistently improves performance across all three tested conditions. This suggests that the benefits of this constraint are not limited to specific datasets, demonstrating its potential for generalization.

\begin{table}[t]
\centering
\setlength{\tabcolsep}{0.16em}
\fontsize{9pt}{12pt}\selectfont
\caption{Impact of LEC Loss across three conditions on Inria 3DMovie dataset.}   
   \begin{tabular}{*{8}{c}}
  \toprule
  \multirow{2}{*}{\textbf{Geo.}} & \multirow{2}{*}{\textbf{View.}} & \multirow{2}{*}{\textbf{LEC Loss}} &\multicolumn{4}{c}{\textbf{Inria 3DMovie}} &  \\
  \cmidrule(lr){4-7} 
  & & & SIoU$\uparrow$ & RMSE$\downarrow$ & PSNR$\uparrow$ & SSIM$\uparrow$ &\\
  \midrule
  
  $\color{red}\checkmark$ &  &  & 0.2836 & 7.47 & 30.66 & 0.693  \\
  $\color{red}\checkmark$ & & $\color{red}\checkmark$ &  0.2949 & 7.46 & 30.68 & 0.693 \\
   & \color{blue}$\checkmark$  &  & 0.3147 & 7.61 & 30.50 & 0.678 & \\
   & \color{blue}$\checkmark$  & \color{blue}$\checkmark$  & 0.3145 & 7.52 & 30.61 & 0.684 \\
   \color{green}$\checkmark$ & \color{green}$\checkmark$ & & 0.3147 & 7.44 & 30.70 & 0.691 \\
   \color{green}$\checkmark$ & \color{green}$\checkmark$ & \color{green}$\checkmark$ & 0.3186 & 7.31 & 30.85 & 0.697 \\
  \bottomrule
\end{tabular}
\label{tab:ablation of lec loss on inria}
\end{table}

\subsection{Ablation Study on Edge Consistency Loss}
When applying the Edge Consistency loss, we conduct experiments to validate the impact of different $\boldsymbol{\alpha}$ values in \cref{eq:lec loss} within a small range. Using the dual-condition diffusion model, we experiment with $\boldsymbol{\alpha}$ values of 0.75, 1, and 1.25, while $\boldsymbol{\alpha}=0$ represents the absence of the edge constraint. As \cref{tab:ablation on alpha} illustrates, applying the edge consistency constraint at varying strengths consistently leads to improvements in SIoU, indicating that the constraint term is not overly sensitive to the specific $\boldsymbol{\alpha}$ value. We offer an additional analysis: when $\boldsymbol{\alpha}$ is 0, all pixels in the image are optimized equally. The edge constraint, in essence, imposes a stricter penalty on regions that genuinely influence


\begin{table}[]
\centering
\setlength{\tabcolsep}{0.25em}
\fontsize{9pt}{10.2pt}\selectfont
\caption{Impact of EC loss across three conditions. EC loss consistently improves performance, with notable gains in SIoU, the metric for perceived stereo quality.}   
\begin{tabular}{*{6}{c}}
  \toprule
 \multirow{2}{*}{\textbf{LEC Loss}} &\multicolumn{4}{c}{\textbf{Mono2Stereo}} &  \\
  \cmidrule(lr){2-5} 
   &SIoU$\uparrow$ & RMSE$\downarrow$ & PSNR$\uparrow$ & SSIM$\uparrow$ &\\
  \midrule
   0 & 0.2588 & 6.90 & 31.35 & 0.721 \\
   0.75& 0.2608 & 6.83 & 31.45 & \textbf{0.725} \\
   1 & \textbf{0.2619} & \textbf{6.82} & \textbf{31.45} & 0.721 \\
   1.25 & 0.2615 & 6.88 & 31.38 & 0.719\\
  \bottomrule

\end{tabular}
\label{tab:ablation on alpha}
\end{table}

\begin{table*}[h]
\centering
\setlength{\tabcolsep}{0.45em}
\fontsize{9pt}{10pt}\selectfont
\caption{Evaluating models across various scenes.} 
   \begin{tabular}{*{13}{c}}
  \toprule
  \multirow{2}{*}{\textbf{Method}} &\multicolumn{2}{c}{\textbf{Indoor}} &\multicolumn{2}{c}{\textbf{Outdoor}} & \multicolumn{2}{c}{\textbf{Complex}} & \multicolumn{2}{c}{\textbf{Simple}} & \multicolumn{2}{c}{\textbf{Animation}}  & \multicolumn{2}{c}{\textbf{Video}} \\
  \cmidrule(lr){2-3} \cmidrule(lr){4-5} \cmidrule(lr){6-7} \cmidrule(lr){8-9}  \cmidrule(lr){10-11}   \cmidrule(lr){12-13} 
   & SIoU$\uparrow$ & RMSE$\downarrow$ & SIoU$\uparrow$ & RMSE$\downarrow$& SIoU$\uparrow$ & RMSE$\downarrow$& SIoU$\uparrow$ & RMSE$\downarrow$& SIoU$\uparrow$ & RMSE$\downarrow$ & SIoU$\uparrow$ & RMSE$\downarrow$   \\
  \midrule

  StereoDiffusion~\cite{wang2024stereodiffusion}  & 0.2387  & 7.48 & 0.2441 & 7.68 & 0.2182 & 7.78 & 0.2571 & 6.17 & 0.2296 & 8.01 & 0.1992 & 8.38\\

  Geometric Condition  & 0.2505 & 5.31 & 0.2543 & 5.74 & 0.2561 & 5.94 & 0.2791 & 4.28 & 0.2525 & 5.73 & 0.2610 & 5.61\\
  Viewpoint Condition & 0.2761 & 5.71 & 0.2824 & 6.02 & 0.2713 & 6.62 & 0.2986 & 5.76 & 0.2764 & 6.49 & 0.2735 & 5.95 \\
   Dual Condition & 0.2819 & 5.21 & 0.2969 & 5.65 & 0.2894 & 5.78 & 0.3095 & 4.29 & 0.2999 & 5.76 &  0.2817 & 5.50\\ 
  \bottomrule
  







\end{tabular}
   
\label{tab:5settings}
\end{table*}

\end{document}